\pdfoutput=1

\documentclass[11pt]{article}

\usepackage{EMNLP2022}

\usepackage{times}
\usepackage{latexsym}
\usepackage{amsfonts}
\usepackage{graphicx}
\usepackage{url} 
\usepackage{amsmath}
\usepackage{ulem}
\usepackage{booktabs}
\usepackage{multirow}
\usepackage{enumitem}
\usepackage[T1]{fontenc}

\usepackage[utf8]{inputenc}

\usepackage{microtype}

\usepackage{inconsolata}

\title{InfoCSE: Information-aggregated Contrastive Learning of Sentence Embeddings}

\author{
    Xing Wu\textsuperscript{\rm 1,2,3},Chaochen Gao\textsuperscript{\rm 1,2}\thanks{The first two authors contribute equally.},Zijia Lin\textsuperscript{\rm 3},Jizhong Han\textsuperscript{\rm 1},Zhongyuan Wang\textsuperscript{\rm 3},Songlin Hu\textsuperscript{\rm 1,2}\thanks{Corresponding author.}
    \\
    \textsuperscript{\rm 1}Institute of Information Engineering, Chinese Academy of Sciences\\
    \textsuperscript{\rm 2}School of Cyber Security, University of Chinese Academy of Sciences\\
    \textsuperscript{\rm 3}Kuaishou Technology
    \\
    \{gaochaochen,zangliangjun,hanjizhong,husonglin\}@iie.ac.cn
    \\\{wuxing,wangzhongyuan\}@kuaishou.com, linzijia07@tsinghua.org.cn
}

\begin{document}
\maketitle
\begin{abstract}
Contrastive learning has been extensively studied in sentence embedding learning, which assumes that the embeddings of different views of the same sentence are closer. The constraint brought by this assumption is weak, and a good sentence representation should also be able to reconstruct the original sentence fragments. Therefore, this paper proposes an \textbf{info}rmation-aggregated \textbf{c}ontrastive learning framework for learning unsupervised \textbf{s}entence \textbf{e}mbeddings, termed InfoCSE.
InfoCSE forces the representation of [CLS] positions to aggregate denser sentence information by introducing an additional Masked language model task and a well-designed network.
We evaluate the proposed InfoCSE on several benchmark datasets w.r.t the semantic text similarity (STS) task.
Experimental results show that InfoCSE outperforms SimCSE by an average Spearman correlation of 2.60\% on BERT-base, and 1.77\% on BERT-large, achieving state-of-the-art results among unsupervised sentence representation learning methods.
Our code are available at \href{https://github.com/caskcsg/sentemb/tree/main/InfoCSE}{github.com/caskcsg/sentemb/tree/main/InfoCSE}.
\end{abstract}

\begin{figure*}
\centering
\includegraphics[width=16cm]{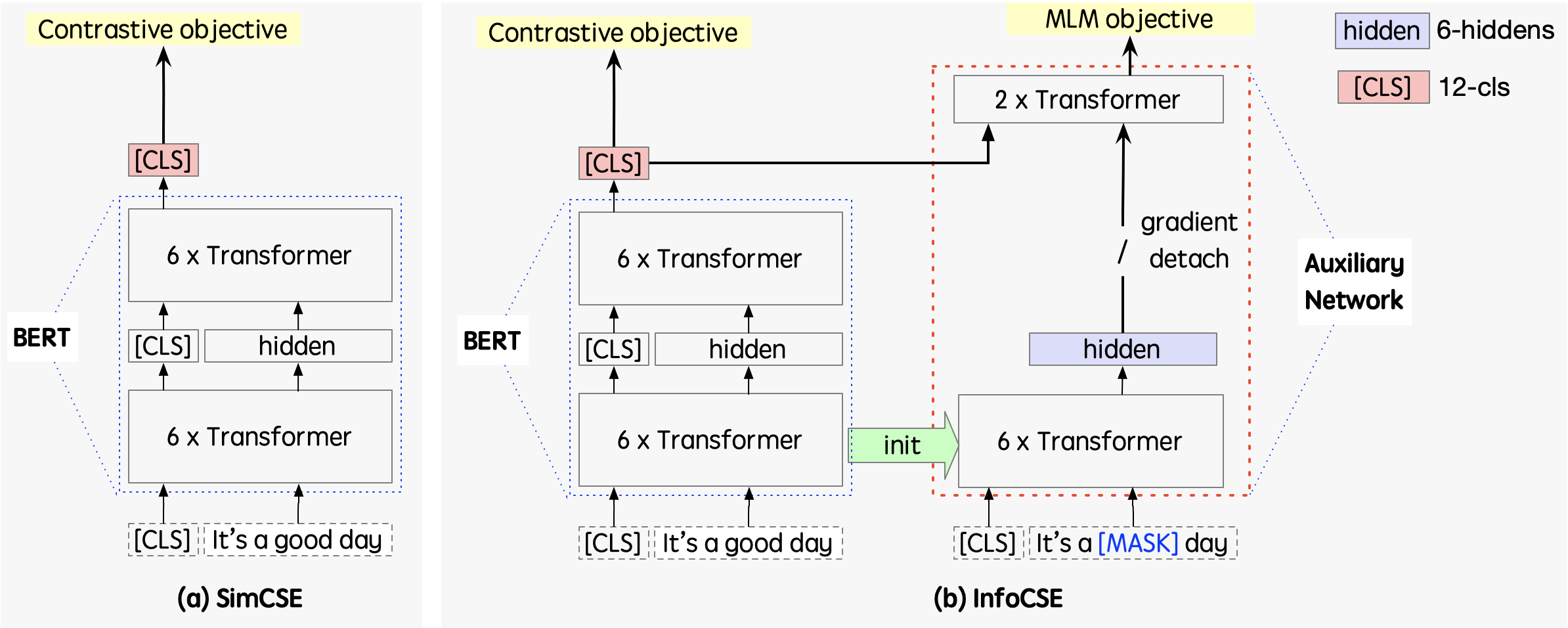}
\caption{Comparison of InfoCSE and SimCSE structures. SimCSE learns sentence representations through contrastive learning on the [CLS] output embeddings of the BERT model. In addition to contrastive learning, InfoCSE designs an auxiliary network for sentence reconstruction with the [CLS] embeddings, enabling to learn better sentence representations.}
\label{InfoCSE_framework}
\end{figure*}

\section{Introduction}
Sentence embeddings aim to capture rich semantic information to be applied in many downstream tasks \cite{zhang2020unsupervised,wu2020clear,liu2021fast}.
Recently, researchers have started to use contrastive learning to learn better unsupervised sentence embeddings \cite{gao2021simcse,yan2021consert, wu2021esimcse, zhou2022debiased, wu2021smoothed, chuang2022diffcse}.
Contrastive learning methods assume that effective sentence embeddings should bring similar sentences closer and dissimilar sentences farther.
These methods use various data augmentation methods to randomly generate different views for each sentence and constrain one sentence semantically to be more similar to its augmented counterpart than any other sentence. 
SimCSE \cite{gao2021simcse} is the representative work of contrastive sentence embedding, which uses dropout acts as minimal data augmentation. 
SimCSE encodes the same sentence twice into embeddings to obtain ``positive pairs" and takes other sentence embeddings in the same mini-batch as ``negatives". There have been many improvements to SimCSE, including enhancing positive and negative sample building methods \cite{wu2021esimcse}, alleviating the influence of improper mini-batch negatives \cite{zhou2022debiased}, and learning sentence representations that are aware of direct surface-level augmentations \cite{chuang2022diffcse}.
\begin{table}[!tbp]
\centering
\small

\setlength{\tabcolsep}{8mm}{\begin{tabular}{lc}
\toprule  
\textbf{Model}& \textbf{STS-B} \\
\midrule
SimCSE-BERT$_{base}$  & 86.2 \\
\midrule
w/ MLM & \\
$\lambda = 0.01$ & 85.7 \\
$\lambda = 0.1$ & 85.7 \\
$\lambda = 1$ & 85.1 \\
\bottomrule
\end{tabular}}
\caption{Table from SimCSE \cite{gao2021simcse}. The masked language model (MLM) objective brings a consistent drop to the SimCSE model in semantic textual similarity tasks. ``w/" means ``with", $\lambda$ is the balance hyperparameter for MLM loss.}
\label{table_mlm_simcse}
\end{table}

Although SimCSE and its variants have achieved good results and can learn sentence embeddings that can distinguish different sentences, this is not enough to indicate sentence embeddings already contain the semantics of sentences well. 
If a sentence embedding is sufficiently equivalent to the semantics of the sentence, it should also be able to reconstruct the original sentence to a large extent \cite{lu2021less}.
However, as shown in Table \ref{table_mlm_simcse} from \cite{gao2021simcse}, experiments show that ``the masked language model objective brings a consistent drop in semantic textual similarity tasks" in SimCSE.
This is due to the gradients of the MLM objective optimization will easily over-update the parameters of the encoder network, thus causing disturbance to the contrastive learning task.
Therefore, it is not an easy job to incorporate sentence reconstruction task in contrastive sentence embedding learning.

To improve contrastive sentence embedding learning with the sentence reconstruction task, we propose an information-aggregated contrastive learning framework, termed InfoCSE. The previous work \cite{gao2021simcse} shares the encoder when jointly optimizing and contrastive learning objective and MLM objective. Unlike \cite{gao2021simcse}, we design an auxiliary network to optimize the MLM objective, as shown in Figure \ref{InfoCSE_framework}-(b). 
The auxiliary network is an 8-layer transformer network consisting of a frozen lower six layers and an additional two layers.
The auxiliary network takes two inputs. One is the sentence embedding of the original text, and the other is the masked text.
The sentence embedding is a vector of the [CLS] position encoded by 12 layers of BERT, abbreviated as \textit{12-cls}.
The lower six Transformer layers encode the masked text to output hidden states at each position. We collectively refer to \text{non}-[CLS] positions' representations as \textit{6-hiddens}.
Then, we feed the concatenation of the \textit{12-cls} and \textit{6-hiddens} into the additional two layers to perform token prediction for the masked positions.
Such a design brings two benefits.
\begin{itemize}[leftmargin=*,itemsep=0pt,topsep=0pt,parsep=0pt]
\item
Since the auxiliary network only contains 8 Transformer layers and the frozen lower six layers cannot be optimized, the sentence reconstruction ability is limited.
Also, the \textit{non}-[CLS] embeddings are the outputs of the 6th Transformer layer, with insufficient semantic information learned. So the MLM task is forced to rely more on the \textit{12-cls} embedding, encouraging the \textit{12-cls} embedding to encode richer semantic information.
\item
The gradient update of the auxiliary network is only back-propagated to the BERT network through the \textit{12-cls} embedding. Compared to perfTorming the MLM task directly on the BERT, the effect of gradient updates using the \textit{12-cls} embedding will be much smaller and will not cause large perturbations to the contrastive learning task.
\end{itemize}
Therefore, under the InfoCSE framework, the \textit{12-cls} embedding learned can be distinguished from other sentence embeddings through contrastive learning and reconstructing sentences through auxiliary MLM training, while avoiding the disadvantage that the gradient of MLM objective will over-update the parameters of the encoder network.

Experiments on the semantic text similarity (STS) tasks show that InfoCSE outperforms SimCSE by an average Spearman correlation of 2.60\%, 1.77\% on the base of BERT-base, BERT-large, respectively.
InfoCSE also significantly outperforms SimCSE on the open-domain information retrieval benchmark BEIR.
We also conduct a set of ablation studies to illustrate the soundness of InfoCSE design.

\section{Backgrounds}
\paragraph{Definitions} Let's define some symbols first. Suppose we have a set of sentences $x \in \mathbb{X}$, and a 12-layer Transformer blocks BERT model $Enc$. For a sentence $ x$ of length $l$, we append a special [CLS] token to it, and then feed it into BERT for encoding. The output of each layer is a vector list of length $l+1$, also called hidden states. 
We use $H$ to denote the last layer's hidden states and $h$ to refer to the vector at [CLS] position.
In addition, we use $M$ to represent the hidden state of the 6th layer, $M^{>0}$ to represent the other hidden states of the 6th layer except the [CLS] position,

\paragraph{SimCSE} As shown in subfigure (a) of Figure \ref{InfoCSE_framework}, we plot the structure diagram of SimCSE-BERT$_{base}$. 
The model's output is the [CLS] position's vector, which is used as the semantic representation of the sentence.
SimCSE uses the same sentence to construct semantically related positive pairs $\langle x, x^{+} \rangle$, i.e. $x^{+} = x$. Specifically, SimCSE uses dropout as the minimum data augmentation, feeding the same input $x$ twice to the encoder with different dropout masks $z$ and $z^{+}$, and outputs the hidden states of the last layer:
\begin{equation}
H=Enc\left(x, z\right), 
H^{+}=Enc\left(x, z^{+}\right)
\end{equation}
A pooler layer $Pooler$ is further applied to the hidden states as follows:
\begin{equation}
h=Pooler\left(H\right), 
h^{+}=Pooler\left(H^{+}\right)
\end{equation}
Then, for a mini-batch $B$,
the contrastive learning objective w.r.t $x$ is formulated as:
\begin{equation}
\mathcal{L}^{\mathrm{cl}}=-\log \frac{\exp(\mathrm{sim}\left(h, h^{+}\right) / \tau)}{\sum\limits_{h^{\prime} \in B} \exp(\mathrm{sim}\left(h,h^{\prime +}\right) / \tau)}
\end{equation}
, where $\tau$ is a temperature hyperparameter and $\mathrm{sim}\left(h, h^{+}\right)$ is  the  similarity metric,  which  is  typically the cosine similarity function.

\paragraph{BERT's MLM objective} MLM randomly masks out a subset of input tokens and requires the model to predict them. Given a sentence $x$, following BERT \cite{devlin2018bert}, we randomly replace 15\% of the tokens with \textit{[MASK]} and get a masked sentence $\widehat{x}$.
Then $\widehat{x}$ will be feed into the BERT, and the hidden states of the last layer $H$ will be projected through a matrix $W$ to predict the original token of each masked position. The process uses the cross entropy loss function $CE$ for optimization:
\begin{equation}
\mathcal{L}^{\mathrm{mlm}}=\sum_{j \in masked} CE\left(H^{j}W, \widehat{x}^{j}\right)
\end{equation}
, where $masked$ denotes the masked positions.
\section{InfoCSE: Information-aggregated Contrastive Learning}
In this section, we first introduce the MLM pre-training of the auxiliary network, and then describe how to jointly train the contrastive learning objective and the MLM objective with the auxiliary network.
\subsection{Pre-training of The Auxiliary Network}
As shown in Figure \ref{InfoCSE_framework}-(b).
the auxiliary network is an 8-layer transformer network consisting of a lower six layers and an additional two layers.
In the pre-training phase of the auxiliary network, the auxiliary network and the BERT network share the lower six layers. 
We optimize two MLM objectives simultaneously using the same input.
The first is BERT's MLM objective, which we have already covered. The second is the MLM objective of the auxiliary network. We concatenate the vector at [CLS] position of BERT's 12th layer ($h$) and the hidden states of the 6th layer except for the [CLS] position ($M^{>0}$):
$$\overset{\sim} H = [h, M^{>0}]$$
Then $\overset{\sim} H$ is fed into the additional two layers, and the output hidden state will be used to calculate the cross-entropy loss:
\begin{equation}
\mathcal{L}^{\mathrm{aux}}=\sum_{j \in masked} CE\left(\overset{\sim}H^{j}W, \widehat{x}^{j}\right)
\end{equation}
Therefore, the pre-training loss of the entire auxiliary network is defined as the sum of the two MLM losses:
\begin{equation}
\mathcal{L}^{\mathrm{pretrain}} = \mathcal{L}^{\mathrm{aux}} + \mathcal{L}^{\mathrm{mlm}}
\end{equation}
The output projection matrix $W$ is shared between the two MLM losses.

\subsection{Joint Training of MLM and Contrastive Learning}
When jointly training the contrastive learning and MLM objectives, the auxiliary network no longer shares the lower six layers with BERT. But they are both initialized with the parameters from the first stage.
While we still optimize the MLM objective for the auxiliary network, the lower six layers' parameters are \textbf{frozen}.
For a sentence $x$, it will be copied twice. One copy is used to form a positive pair $\langle x, x^{+} \rangle$ for contrastive learning. The other is used to generate the masked input $\widehat{x}$, which will be used in the MLM task.
The loss of contrastive learning is the same as SimCSE, which we have already introduced.
The MLM loss of the auxiliary network has changed a little, here the $M^{>0}$ is the output from the lower six layers of the auxiliary network:
$$C = [h, Detach(M^{>0})]$$,
where $Detach$ means that the gradient will not back-propagate through $M^{>0}$.
The loss for joint training is defined as the sum of the contrastive learning loss and the auxiliary network's MLM loss:
\begin{equation}
\mathcal{L}^{\mathrm{joint}} = \mathcal{L}^{\mathrm{cl}} + \mathcal{L}^{\mathrm{aux}} * \lambda
\end{equation}
, where $\lambda$ is the balanced hyperparameter for the MLM loss of the auxiliary network.

\section{Experiment}
\subsection{Setup}
When pretraining the auxiliary network, following \cite{gao2021condenser} we use Bookcorpus \cite{zhu2015aligning} and Wikipedia\footnote{https://dumps.wikimedia.org/enwiki/latest/enwiki-latest-pages-articles.xml.bz2} as our datasets.  
When jointly optimizing contrastive objective and the auxiliary MLM objective, for a fair comparison, our experimental setup mainly follows SimCSE. We use 1-million sentences randomly drawn from English Wikipedia for training\footnote{https://huggingface.co/datasets/princeton-nlp/datasets-for-simcse/resolve/main/wiki1m\_for\_simcse.txt}.
All evaluations are directly based on the sentence embeddings output by the BERT network, and the auxiliary network will not be used.

\paragraph{Semantic Textual Similarity Tasks} The semantic textual similarity task measures the capability of sentence embeddings, and we conduct our experiments on seven standard semantic textual similarity (STS) datasets. 
STS12-STS16 datasets \cite{agirre2012semeval,agirre2013sem,agirre2014semeval,agirre2015semeval,agirre2016semeval} do not have train or development sets, and thus we evaluate the models on the development set of STS-B \cite{cer2017semeval} to search for better settings of the hyper-parameters.
The SentEval toolkit\footnote{https://github.com/facebookresearch/SentEval} is used for evaluation, and Spearman correlation coefficient \footnote{\url{https://en.wikipedia.org/wiki/Spearman\%27s_rank_correlation_coefficient}}  is used to report the model performance. 
\paragraph{Open-domain Retrieval Tasks} 
 Sentence embeddings are also commonly used on the retrieval tasks, so we evaluate the zero-shot performance of embeddings on the information retrieval benchmark BEIR \cite{thakur2021beir}.
BEIR contains 9 open-domain information retrieval tasks (fact checking, citation prediction, duplicate question retrieval, parameter retrieval, news retrieval, question answering, tweet retrieval, biomedical IR, entity retrieval) from 18 different datasets. We evaluate InfoCSE on the 14 publicly available datasets. 

\subsection{Training Details}
We start from the publicly pre-trained checkpoint of BERT-base or BERT-large.
The hyperparameters basically follow Condenser \cite{gao2021condenser} and SimCSE \cite{gao2021simcse}.
When pretraining the auxiliary network, following \cite{gao2021condenser}, we train 8 epochs using the Adam optimizer with learning rate $= 1e-4$, global batch size $= 1024$ on 8 Nvidia V100 GPUs \footnote{The pre-training stage adopts 8 training epochs without additional convergence criteria, so no extra validation set is required.}. The hyperparameter $\lambda^{\mathrm{mlm}} = 1.0$ to balance the two MLM losses.
When jointly optimizing contrastive objective and the auxiliary MLM objective, following \cite{gao2021simcse}, we train one epoch using the Adam optimizer with learning rate $= 3e-5$, batch size $= 64$ on a single Nvidia 3090 GPU. The hyperparameter $\lambda^{\mathrm{mlm}} = 0.005$ to balance the MLM loss and contrastive learning loss. We evaluate the model every 125 training steps on the development set of STS-B and keep the best checkpoint for the final evaluation on test sets.

\subsection{Main Results}
\begin{table*}[!ht]
\centering
\small
\begin{tabular}{lcccccccl}
\toprule  
\textbf{Model}& \textbf{STS12} & \textbf{STS13} & \textbf{STS14} & \textbf{SICK15} & \textbf{STS16} & \textbf{STS-B} & \textbf{SICK-R} & \textbf{Avg.} \\
\midrule
\midrule
GloVe embeddings(avg.) & 55.14 & 70.66 & 59.73 & 68.25 & 63.66 & 58.02 & 53.76 & 61.32 \\
BERT$_{base}$(first-last avg.) & 39.70 & 59.38 & 49.67 & 66.03 & 66.19 & 53.87 & 62.06 & 56.70 \\
BERT$_{base}$-flow & 58.40 & 67.10 & 60.85 & 75.16 & 71.22 & 68.66 & 64.47 & 66.55 \\
BERT$_{base}$-whitening & 57.83 & 66.90 & 60.90 & 75.08 & 71.31 & 68.24 & 63.73 & 66.28 \\
IS-BERT$_{base}$ $\triangle$ & 56.77 & 69.24 & 61.21 & 75.23 & 70.16 & 69.21 & 64.25 & 66.58 \\
CT-BERT$_{base}$ $\triangle$ & 61.63 & 76.80 & 68.47 & 77.50 & 76.48 & 74.31 & 69.19 & 72.05 \\
ConSERT$_{base}$ $\heartsuit$ & 64.64 & 78.49 & 69.07 & 79.72 & 75.95 & 73.97 & 67.31 & 72.74 \\
BERT$_{base}$-flow$\diamondsuit$ & 63.48 & 72.14 & 68.42 & 73.77 & 75.37 & 70.72 & 63.11 & 69.57 \\
SG-OPT-BERT$_{base}$ $\spadesuit$ & 66.84	& 80.13	& 71.23	& 81.56	& 77.17  & 77.23 & 68.16 & 74.62 \\
Mirror-BERT$_{base}$ $\sharp$ & 69.10 & 81.10 &  73.00 & 81.90 & 75.70 & 78.00 & 69.10 & 75.40 \\
SimCSE-BERT$_{base}$ $\clubsuit$ & 68.40 & 82.41 & 74.38 & 80.91 & 78.56 & 76.85 & 72.23 & 76.25 \\
ESimCSE-BERT$_{base}$ $\star$ & \textbf{73.40} & 83.27 & \textbf{77.25} & 82.66 & 78.81 & 80.17 & \textbf{72.30} & 78.27 \\
DiffCSE-BERT$_{base}$ $\P$ & 72.28 & 84.43 & 76.47 & 83.90 & 80.54 & 80.59 & 71.23 & 78.49 \\
InfoCSE-BERT$_{base}$ & 70.53 & \textbf{84.59} & 76.40 & \textbf{85.10} & \textbf{81.95} & \textbf{82.00} & 71.37 & \textbf{78.85} \\
\midrule 
ConSERT$_{large}$ $\heartsuit$ & 70.69 & 82.96 & 74.13 & 82.78 & 76.66 & 77.53 & 70.37 & 76.45 \\
BERT$_{large}$-flow$\diamondsuit$ & 65.20 & 73.39 & 69.42 & 74.92 & 77.63 & 72.26 & 62.50 & 70.76 \\
SG-OPT-BERT$_{large}$ $\spadesuit$ & 67.02 & 79.42 & 70.38 & 81.72 & 76.35 & 76.16 & 70.20 & 74.46  \\
SimCSE-BERT$_{large}$ $\clubsuit$ & 70.88 & 84.16 & 76.43 & 84.50 & 79.76 & 79.26 & 73.88 & 78.41 \\
ESimCSE-BERT$_{large}$ $\star$ & \textbf{73.21} & 85.37 & \textbf{77.73} & 84.30 & 78.92 & 80.73 & \textbf{74.89} & 79.31 \\
DiffCSE-BERT$_{large}$ $\dagger$  & 72.11 & 84.99 & 76.19 & 85.09 & 78.65 & 80.34 & 73.93 & 78.76 \\ 
InfoCSE-BERT$_{large}$ & 71.89 & \textbf{86.17} & 77.72 & \textbf{86.20} & \textbf{81.29} & \textbf{83.16} & 74.84 & \textbf{80.18} \\
\bottomrule
\end{tabular}
\caption{Sentence embedding performance on 7 semantic textual similarity (STS) test sets. $\clubsuit$ : results from official published model by ~\cite{gao2021simcse}.$\heartsuit$ : results from ~\cite{yan2021consert}. $\spadesuit$ : results from ~\cite{kim2021self}. $\diamondsuit$ : results from ~\cite{li2020sentence}. $\triangle$ : results are reproduced and reevaluated by ~\cite{gao2021simcse}. $\sharp$ : results from ~\cite{liu2021fast}. $\star$ : results from ~\cite{wu2021esimcse}. $\P$ : results from ~\cite{chuang2022diffcse}. $\dagger$: The original paper does not report the results of BERT-large, so we use the official public code to perform a grid search on important hyperparameters for the best results.}
\label{table_test_best}
\end{table*}

\paragraph{Baselines}
We compare our model with many strong unsupervised baselines including SimCSE \cite{gao2021simcse}, IS-BERT \cite{zhang2020unsupervised}, CT-BERT \cite{carlsson2021semantic}, ConSERT \cite{yan2021consert}, SG-OPT \cite{kim2021self}, Mirror-BERT \cite{liu2021fast}, ESimCSE \cite{wu2021esimcse}, DiffCSE \cite{chuang2022diffcse} and some post-processing methods like BERT-flow \cite{li2020sentence} and BERT-whitening \cite{su2021whitening} along with some naive baselines like averaged GloVe embeddings \cite{pennington2014glove} and averaged first and last layer BERT \cite{devlin2018bert} embeddings. Some baseline methods are evaluated on both the base and large parameter scales, while others are only evaluated on the base parameter scale.

\paragraph{Semantic Textual Similarity (STS)}
Table \ref{table_test_best} shows different methods' performances on seven semantic textual similarity (STS) test sets. It can be seen that InfoCSE improves the measurement of semantic textual similarity in different parameter scale settings over previous methods. Specifically, InfoCSE-BERT$_{base}$ outperforms SimCSE-BERT$_{base}$ by +2.60\%, InfoCSE-BERT$_{large}$ outperforms SimCSE-BERT$_{large}$ by +1.77\%. Compared with other recent improvements to SimCSE, InfoCSE also achieves better results.

\paragraph{Open-domain Retrieval Task}
Table \ref{tabler_zero} shows the zero-shot performance of different methods on the BEIR benchmark. On the average performance of 14 datasets, InfoCSE outperforms other methods substantially, and InfoCSE achieves the best performance on 10 datasets. Although the recent improvements to SimCSE have achieved better results than SimCSE on STS tasks, their performance on open-domain information retrieval is similar to SimCSE. In contrast, InfoCSE significantly improves on both the STS and open-domain information retrieval tasks over SimCSE, which shows that InfoCSE has better generalization ability.
\begin{table*}[!ht]
\centering
\begin{tabular}{l|cc|cc|cc|cc}
\toprule  
\textbf{Dataset}& \multicolumn{2}{c|}{SimCSE}& \multicolumn{2}{c|}{ESimCSE} & \multicolumn{2}{c|}{DiffCSE} & \multicolumn{2}{c}{InfoCSE} \\
& base & large & base & large & base & large & base & large \\
\midrule 
\textbf{trec-covid} & 0.2750  & 0.2264  & 0.2291  & 0.2829  & 0.2368  & 0.2291  & \textbf{0.3937}  & \underline{0.3166} \\
\textbf{nfcorpus} & 0.1048  & 0.1356  & 0.1149  & 0.1483  & 0.1204  & 0.1470  & 0.1358  & \textbf{0.1576} \\
\textbf{nq} & 0.1628  & 0.1671  & 0.0935  & 0.1705  & 0.1188  & 0.1556  & \textbf{0.2023}  & \underline{0.1790} \\
\textbf{fiqa} & 0.0985  & 0.0975  & 0.0731  & \textbf{0.1117}  & 0.0924  & \underline{0.1027}  & 0.0991  & 0.1000 \\
\textbf{arguana} & 0.2796  & 0.2078  & \underline{0.3376}  & 0.2604  & 0.2500  & 0.2572  & 0.3244  & \textbf{0.4133} \\
\textbf{webis-touche2020} & \textbf{0.1342}  & 0.0878  & 0.0786  & 0.1057  & 0.0912  & 0.0781  & \underline{0.0935}  & 0.0920 \\
\textbf{quora} & 0.7375  & 0.7511  & 0.7411  & 0.7615  & 0.7491  & 0.7471  & \underline{0.8241}  & \textbf{0.8268} \\
\textbf{cqadupstack} & 0.1349  & 0.1082  & 0.1276  & 0.1196  & 0.1197  & 0.1160  & \textbf{0.2097}  & \underline{0.1881} \\
\textbf{dbpedia-entity} & 0.1662  & 0.1495  & 0.1260  & 0.1650  & 0.1537  & 0.1571  & \textbf{0.2101}  & \underline{0.1838} \\
\textbf{scidocs} & 0.0611  & 0.0688  & 0.0657  & 0.0796  & 0.0673  & 0.0699  & \underline{0.0837}  & \textbf{0.0859} \\
\textbf{climate-fever} & \textbf{0.1420}  & 0.1065  & 0.0796  & 0.1302  & 0.1019  & \underline{0.1087}  & 0.0937  & 0.0840 \\
\textbf{scifact} & 0.2492  & 0.2541  & 0.3013  & 0.2875  & 0.2666  & 0.2811  & \underline{0.3269}  & \textbf{0.3801} \\
\textbf{hotpotqa} & 0.2382  & 0.1896  & 0.1213  & 0.1970  & 0.1730  & 0.2068  & \textbf{0.3177}  & \underline{0.2781} \\
\textbf{fever} & \textbf{0.2916}  & 0.1776  & 0.0756  & 0.1689  & 0.1416  & 0.1849  & \textbf{0.1978}  & 0.1252 \\
\midrule
\textbf{average} & 0.2197  & 0.1948  & 0.1832  & 0.2135  & 0.1916  & 0.2030  & \textbf{0.2509}  & \underline{0.2436} \\
\bottomrule
\end{tabular}
\caption{Zero-shot evaluation results on the BEIR benchmark. All scores denote \textbf{nDCG@10}. The best score on a given dataset is marked in \textbf{bold}, and the second best is \underline{underlined}.}
\label{tabler_zero}
\end{table*}

\section{Ablation Studies}
We perform an extensive series of ablation studies to InfoCSE with BERT$_{base}$ scale on the development set of STS-B.

\subsection{The Impact of Auxilary Network}
\paragraph{Pre-training of The Auxiliary Network}
We compared the effect of directly using the parameters of the lower six layers and the upper two layers of BERT as the auxiliary network without pre-training.
As shown in Table \ref{tablempact_aux_pretrain}, the joint training effect using the auxiliary network without pre-training is significantly worse, illustrating the importance of the pre-training process.
Pre-training enables the additional two-layer transformer to fit the MLM task well, avoiding large gradient oscillations during the joint learning.
\paragraph{Joint Training of MLM and Contrastive Learning}
We remove different loss terms to study their importance in joint training. We list the results in Table \ref{table_diff_losses}. When we remove the MLM loss, the InfoCSE model degenerates the SimCSE model, and the performance on STS-B drops by 3\%. When we remove the contrastive loss, the InfoCSE model degenerates the pre-trained auxiliary network, the performance drops sharply. This result shows that both the contrastive loss and the auxiliary MLM loss are crucial in InfoCSE.

\subsection{The Impact of Gradient Detach in Joint Training}
During joint training, we freeze the parameters of the lower six layers of the auxiliary network through the gradient detach operation. In other words, the lower six layers are only used as feature extractors for the non-[CLS] positions. The additional two transformer layers have a limited number of parameters, and the modeling ability is weak. 
As shown in Table \ref{tablempact_grad_detach}, the joint training effect without gradient separation operation is reduced by 1\%.
Therefore, the gradient detach operation forces the MLM task to rely more on the sentence embedding of \textit{12-cls} embedding, which is beneficial to the learning of sentence embedding. 

\begin{table}[!t]
\centering
\begin{tabular}{lc}
\toprule  
\textbf{Model}& \textbf{STS-B} \\
\midrule
InfoCSE & \textbf{85.49} \\
\textit{w/o pre-training}  & 83.73 \\
\bottomrule
\end{tabular}
\caption{Development set results of STS-B for InfoCSE with or without auxiliary network pre-training. ``w/o" denotes without.}
\label{tablempact_aux_pretrain}
\end{table}

\begin{table}[!t]
\centering
\begin{tabular}{lc}
\toprule  
\textbf{Model}& \textbf{STS-B} \\
\midrule
InfoCSE & \textbf{85.49} \\
\textit{w/o MLM loss}  & 82.45 \\
\textit{w/o Contrastive loss} & 40.00  \\
\bottomrule
\end{tabular}
\caption{Development set results of STS-B for InfoCSE variants, where we vary the objective. ``w/o" denotes without.}
\label{table_diff_losses}
\end{table}

\begin{table}[!t]
\centering
\begin{tabular}{lc}
\toprule  
\textbf{Model}& \textbf{STS-B} \\
\midrule
InfoCSE & \textbf{85.49} \\
\textit{w/o gradient detach}  & 84.41 \\
\bottomrule
\end{tabular}
\caption{Development set results of STS-B for InfoCSE with or without gradient detach. ``w/o" denotes without.}
\label{tablempact_grad_detach}
\end{table}

\begin{table}[!tpb]
\centering
\begin{tabular}{lccccc}
\toprule 
\textbf{Mask Rate} & \textit{10\%} & \textit{15\%} & \textit{20\%}  & \textit{25\%} \\
\textbf{STS-B} & 83.97 & 84.62 & 84.74 & 85.08 \\
\midrule
\textbf{Mask Rate} & \textit{30\%}  & \textit{35\%} & \textit{40\%} & \textit{45\%} \\
\textbf{STS-B} & 84.19 & 84.70 & \textbf{85.49} & 84.13 \\
\bottomrule
\end{tabular}
\caption{Development set results of STS-B when we vary the mask rate.}
\label{table_mlm_rate}
\end{table}

\begin{table}[!tpb]
\centering
\begin{tabular}{lccccc}
\toprule 
$\lambda$ & 0.& 5e-6 & 1e-5  & 5e-5 \\
STS-B & 82.45 & 84.50 & \textbf{85.49} & 84.7 \\
\midrule
$\lambda$ & 1e-4  & 5e-3 & 1e-2 & 5e-2\\
STS-B &  84.29 & 80.48 & 76.15 & 75.27 \\
\bottomrule
\end{tabular}
\caption{Development set results of STS-B when we vary the coefficient $\lambda$.}
\label{table_lambda}
\end{table}

\begin{table}[!t]
\centering
\begin{tabular}{lcc}
\toprule  
\textbf{Model}& \textbf{cls} & \textbf{cls\_before\_pooler} \\
\midrule
SimCSE & 81.72 & 82.45 \\
DiffCSE & 83.90 & 84.56 \\
\midrule
InfoCSE & \textbf{85.08} & \textbf{85.49} \\
\bottomrule
\end{tabular}
\caption{Development set results of STS-B where we vary the pooler choice. ``cls'' denotes using the representation of [CLS] token; ``cls\_before\_pooler'' denotes using the representation of [CLS] token without the extra linear+activation.}
\label{table_pooler}
\end{table}

\subsection{The Impact of Mask Rate}
\cite{wettig2022should} shows that 15\% may not always be the optimal mask rate. Therefore, we further explore the impact of different mask rates for MLM task in joint training. We vary the mask rate from $\textit{10\%}$ to $\textit{45\%}$. As shown in \ref{table_mlm_rate}, when the mask rate increase from $\textit{10\%}$ to $\textit{25\%}$, the model performance steadily increases. When the mask rate is over $\textit{25\%}$, the model performance begins to fluctuate, and the optimal mask rate is $\textit{40\%}$.

\begin{table*}[!htbp]
\centering
\begin{tabular}{lcccccccl}
\toprule 
\textbf{Model} & \textbf{MR} & \textbf{CR} & \textbf{SUBJ} & \textbf{MPQA} & \textbf{SST} & \textbf{TREC} & \textbf{MRPC} & \textbf{Avg.} \\
GloVe embeddings (avg.) & 77.25 & 78.30 & 91.17 & 87.85 & 80.18 & 83.00 & 72.87 & 81.52 \\
Skip-thought & 76.50 & 80.10 & 93.60 & 87.10 & 82.00 & 92.20 & 73.00 & 83.50 \\
\midrule
Avg. BERT embeddings & 78.66 & 86.25 & 94.37 & 88.66 & 84.40 & 92.80 & 69.54 & 84.94 \\
BERT-[CLS] embedding & 78.68 & 84.85 & 94.21 & 88.23 & 84.13 & 91.40 & 71.13 & 84.66 \\
IS-BERT$_{base}$ $\triangle$ & 81.09 & 87.18 & 94.96 & 88.75 & 85.96 & 88.64 & 74.24 & 85.83 \\
SimCSE $\clubsuit$ & 81.18 & 86.46 & 94.45 & 88.88 & 85.50 & 89.80 & 74.43 & 85.81 \\
w/MLM $\clubsuit$ & 82.92 & 87.23 & 95.71 & 88.73 & 86.81 & 87.01 & 78.07 & 86.64 \\
InfoCSE & 81.76 & 86.57 & 94.90 & 88.86 & 87.15 & 90.60 & 76.58 & 86.63 \\
\bottomrule
\end{tabular}
\caption{Results on transfer tasks of different sentence embedding models, in terms of accuracy. $\clubsuit$ : results from official published model by ~\cite{gao2021simcse}.$\heartsuit$ : results from ~\cite{yan2021consert}. $\triangle$ : results are reproduced and reevaluated by ~\cite{gao2021simcse}. $\P$ : results from ~\cite{chuang2022diffcse}.}
\label{table_transfer}
\end{table*}

\subsection{The Impact of Coefficient $\lambda$}
Section 3.2 uses the $\lambda$ coefficient to weight the auxiliary MLM loss and add it with contrastive loss. Because the contrastive learning objective is a relatively easier task, the scale of contrastive loss will be thousands of times smaller than MLM loss. As a result, we need a smaller $\lambda$ to balance these two loss terms. In the Table \ref{table_lambda} we show the STS-B result under different $\lambda$ values. Note that when $\lambda$ goes to zero, the model becomes a SimCSE model. We find that using $\lambda$ = 1e-5 can give us the best performance.

\subsection{The Impact of Pooler}
There are two different pooler methods in SimCSE. One is ``cls\_before\_pooler'', which is the [CLS] representation of BERT's last layer. 
The other is ``cls'', which is ``cls\_before\_pooler'' with additional linear and activation.  Here we compare the effects of using two poolers in different models. 
As shown in Table \ref{table_pooler}, ``cls\_before\_pooler'' is always better than ``cls'', and InfoCSE is significantly better than other models when using either pooler.

\section{Analysis}
In this section, we further analyze the performance of InfoCSE on transfer tasks and in-domain information retrieval task. We also explore whether different auxiliary objectives can coexist.

\subsection{Transfer Tasks}
\paragraph{Dataset} The transfer tasks include: MR (movie review) \cite{pang2005seeing}, CR (product review) \cite{hu2004mining}, SUBJ (subjectivity status) \cite{ pang2004sentimental} , MPQA (opinion-polarity) \cite{wiebe2005annotating}, SST-2 (binary sentiment analysis) \cite{socher2013recursive}, TREC (question-type classification) \cite{voorhees2000building} and MRPC (paraphrase detection) \cite{dolan2005automatically}.
In these transfer tasks, we will use a logistic regression classifier trained on top of the frozen sentence embeddings, following the standard setup \footnote{\url{https://github.com/facebookresearch/SentEval} }.
We show the results of transfer tasks in Table \ref{table_transfer}. 
Compared with SimCSE, InfoCSE can improve the average scores from 85.81\% to 86.63\%. In SimCSE, the authors also propose directly using BERT's MLM task to further boost the performance of transfer tasks, but it brings a consistent drop in STS tasks. Compared with SimCSE with MLM, InfoCSE achieves comparable results on transfer tasks without a drop in STS tasks. This shows that the proposed auxiliary MLM network can simultaneously improve contrastive sentence embedding's performance on semantic text similarity tasks and downstream tasks.

\subsection{In-domain Retrieval Task}
Following \cite{chuang2022diffcse}, we further explore the performance of InfoCSE on the in-domain retrieval task. STS-B test is used in the in-domain retrieval corpus, with 2758 sentences. 
Among them, there are 97 positive pairs $\langle s_1, s_2 \rangle$, and we use all $s1$ to form the query set.
Then we use each query $s_1$ to retrieve its nearest neighbours from the corpus in the sentence embedding space and see whether its corresponding $s_2$ is ranked in the top-1/5/10 order.
We show quantitative results in Table \ref{tablen_retrieval}. InfoCSE outperforms SimCSE and DiffCSE on recall@1/5/10 results, demonstrating InfoCSE's effectiveness for in-domain retrieval tasks.
\begin{table}[!t]
\centering
\begin{tabular}{l|ccc}
\toprule  
\textbf{Model}& \textbf{R@1} & \textbf{R@5} & \textbf{R@10} \\
\midrule
SimCSE & 74.23 & 94.85 & 95.88 \\
DiffCSE & 79.38 & 96.91 & 98.97 \\
\midrule
InfoCSE & \textbf{80.41} & \textbf{100.00} & \textbf{100.00} \\
\bottomrule
\end{tabular}
\caption{In-domain retrieval results.``R@'' denotes recall.}
\label{tablen_retrieval}
\end{table}

\subsection{Compatibility of Different Auxiliary Objectives}
Likewise, DiffCSE also introduces an auxiliary objective, replaced token detection (RTD), which brings non-trivial improvements to the ability of sentence embeddings. The difference is that RTD is a discriminative objective, while MLM is a generative objective. Therefore, we further explore whether these two different auxiliary objectives can coexist. Specifically, we simultaneously apply the [CLS] sentence embedding to the auxiliary MLM objective and RTD objective for joint optimization. As shown in Table \ref{table_combine}, it is encouraging that the two objectives are well compatible and the combined approach achieves further improvements. This suggests that designing auxiliary objectives could be a promising direction for improving sentence embedding ability, and different auxiliary objectives have the potential to be compatible with each other if properly designed.
\begin{table}[!t]
\centering
\begin{tabular}{l|ccc}
\toprule  
\textbf{Model}& \textbf{STSB} & \textbf{Avg.} \\
\midrule
SimCSE & 82.45 & 76.25 \\
w/ RTD (DiffCSE) & 84.56 & 78.27 \\
w/ MLM (InfoCSE) & 85.49 & 78.49 \\
\midrule
w/ RTD + MLM & \textbf{85.83} & \textbf{79.39} \\
\bottomrule
\end{tabular}
\caption{The comparison of the improvement brought by different auxiliary objectives to SimCSE. ``w/'' denotes without. ``STS-B'' denotes the best result on the STS-B development set. ``Avg.'' denotes the corresponding average result on 7 semantic textual similarity (STS) test sets.}
\label{table_combine}
\end{table}



\section{Related Work}
Unsupervised sentence representation learning has been widely studied. ~\cite{socher2011dynamic,hill2016learning,le2014distributed} propose to learn sentence representation according to the internal structure of each sentence. ~\cite{kiros2015skip,logeswaran2018efficient} predict the surrounding sentences of a given sentence based on the distribution hypothesis. ~\cite{pagliardini2017unsupervised} propose Sent2Vec, a simple unsupervised model allowing to compose sentence embeddings using word vectors along with n-gram embeddings. Recently, contrastive learning has been explored in unsupervised sentence representation learning and has become a promising trend \cite{zhang2020unsupervised,wu2020clear,meng2021coco,liu2021fast,gao2021simcse,yan2021consert, wu2021esimcse, zhou2022debiased, wu2021smoothed, chuang2022diffcse}. Those contrastive learning based methods for sentence embeddings are generally based on the assumption that a good semantic representation should be able to bring similar sentences closer while pushing away dissimilar ones.
The most related ones are \cite{gao2021simcse} and \cite{chuang2022diffcse}.We solve the problem in \cite{gao2021simcse} that the MLM task will reduce the model effect on STS through designing an auxiliary network. Moreover,  we achieve further improvement on STS via information aggregation. \cite{chuang2022diffcse} is similar to us in network structure, and we both design an auxiliary network. But the motivations are quite different. \cite{chuang2022diffcse} aims to introduce equivariant contrastive learning to SimCSE.  Then the sentence representations will be aware of, but not necessarily invariant to, direct surface-level augmentations. We aim to simultaneously achieve information aggregation while solving the compatibility problem of MLM and contrastive learning.

\section{Conclusion and Future Work}
In this paper, we present InfoCSE, an information-aggregated contrastive learning framework for learning unsupervised sentence embeddings. Empirical improvements on different datasets show the effectiveness and transferability of InfoCSE. We also conduct extensive ablation studies to demonstrate the different modeling choices in InfoCSE.
We believe that improving sentence embeddings by optimizing auxiliary objectives is a very promising direction. In the future, we will further explore its potential on more tasks, such as dense passage retrieval \cite{wu2022contextual}.

\section{Limitations}
One limitation of our work is that we do not explore the supervised setting that uses human-labeled NLI datasets to further boost the performance.

\bibliography{anthology,custom}
\bibliographystyle{acl_natbib}

\end{document}